\documentclass[10pt,twocolumn,letterpaper]{article}

\usepackage[pagenumbers]{cvpr} 

\usepackage[dvipsnames]{xcolor}

\usepackage{mathtools}
\usepackage{amsfonts}
\usepackage{amssymb}
\usepackage{dblfloatfix}
\usepackage{appendix}
\usepackage{multirow}
\usepackage{multicol}

\newcommand{\subsectionnonum}[1]{\vspace{1pt}\noindent\textbf{#1}}

\definecolor{cvprblue}{rgb}{0.21,0.49,0.74}
\usepackage[pagebackref,breaklinks,colorlinks,citecolor=cvprblue]{hyperref}

\title{Neural Radiance and Gaze Fields for Visual Attention Modeling \\in 3D Environments}

\author{
Andrei Chubarau, Yinan Wang, James J. Clark\\
McGill University\\
Centre for Intelligent Machines\\
{\tt\small \{andrei.chubarau@, yinan.wang2@\}mail.mcgill.ca, james.clark1@mcgill.ca}
}

\begin{document}
\maketitle
\begin{abstract}
We introduce Neural Radiance and Gaze Fields (NeRGs), a novel approach for representing visual attention in complex environments. Much like how Neural Radiance Fields (NeRFs) perform novel view synthesis, NeRGs reconstruct gaze patterns from arbitrary viewpoints, implicitly mapping visual attention to 3D surfaces. We achieve this by augmenting a standard NeRF with an additional network that models local egocentric gaze probability density, conditioned on scene geometry and observer position. The output of a NeRG is a rendered view of the scene alongside a pixel-wise salience map representing the conditional probability that a given observer fixates on visible surfaces. Unlike prior methods, our system is lightweight and enables visualization of gaze fields at interactive framerates. Moreover, NeRGs allow the observer perspective to be decoupled from the rendering camera and correctly account for gaze occlusion due to intervening geometry. We demonstrate the effectiveness of NeRGs using head pose from skeleton tracking as a proxy for gaze, employing our proposed gaze probes to aggregate noisy rays into robust probability density targets for supervision.
\end{abstract}

\section{Introduction}
\label{sec:intro}

Visual attention is the selective allocation of focus by viewers to specific elements within a scene, commonly quantified by eye fixation probability. Most prior work addresses attention in 2D images and video, including visual salience in static images \cite{itti1998model, kummerer2014deep, kummerer2016deepgaze, linardos2021deepgaze, kummerer2022deepgaze} and egocentric gaze estimation \cite{Li_2013_ICCV, zhang2017deep, lai2023eye, Wang2024eyeTracking}.  
However, because 2D models are inherently uninformed by object geometry, they often fail to generalize to spatially complex 3D scenes \cite{song20233d}, where depth and occlusion influence attention and vary with the observer’s pose. Challenges in measuring and mapping eye fixations to 3D surfaces further complicate the visualization of gaze patterns in 3D environments \cite{elmadjian20183d, wang2018slam, yang2021visual}.

Modeling visual attention and gaze in 3D requires two main components: (i) a 3D representation of the scene’s geometry describing the shape and position of objects, and (ii) a method for acquiring, representing, and mapping attention data relative to 3D objects and surfaces. Meshes are a common choice for representing 3D objects, and mesh saliency methods \cite{lee2005mesh, song2019, mesh3DSaliency2024, attention3Dshapes} directly predict eye fixation patterns on 3D surfaces for both individual objects and composite scenes. However, these methods are often computationally expensive and depend on the availability of existing meshes. When such data is unavailable, manual modeling is tedious and time-consuming \cite{blender}, whereas reconstructing meshes from 2D images remains challenging \cite{xu2024instantmesh}, particularly for complex or detailed surfaces.

Neural Radiance Fields (NeRFs) \cite{mildenhall2021nerf} provide a powerful alternative for representing 3D scenes, enabling novel view synthesis and faithful reconstruction from arbitrary viewpoints.
Unlike traditional mesh-based approaches that require labor-intensive modeling and texturing, NeRFs can generate high-quality 3D representations within minutes from a set of 2D views of a scene \cite{mueller2022instantNgp}. This makes them particularly well-suited for 3D applications where meshes are unavailable or impractical to produce.
Recent NeRF-based methods also capture 3D object boundaries and semantics, enabling per-object segmentation and labeling \cite{instanceNerf2023, semanticNerf2021, semanticRay2023}. That said, these approaches target static, view-invariant labels, but human visual attention depends on the observer’s visual field, which motivates models that couple 3D geometry with view-dependent perception.

In this work, we address the challenge of visualizing gaze patterns within 3D environments represented by a pre-trained NeRF. We adopt a broad definition of ``gaze'' that encompasses signals derived from salience maps, eye fixations, head pose tracking, or other sources. Framing gaze visualization as a rendering problem, we extend NeRFs to model spatial gaze alongside color and density. This approach enables the synthesis of gaze fields from arbitrary observer positions and offers the flexibility to decouple the observer’s viewpoint from the rendering camera. Because surfaces visible to the rendering camera may be occluded from the perspective of a decoupled observer, modeling gaze occlusion is a critical component of our work. Our combined method builds on efficient NeRF implementations \cite{mueller2022instantNgp}, ensuring real-time interaction and scalability to complex scenes.
Our main contributions are as follows:
\begin{itemize}
\item We introduce Neural Radiance and Gaze Fields (NeRGs), which extend standard NeRFs with view-dependent 3D gaze prediction constrained by scene geometry, implicitly mapping visual attention to 3D surfaces.
\item We train NeRGs using our proposed \emph{gaze probes}, which aggregate local gaze rays into egocentric gaze density distributions anchored in NeRF space, providing a robust supervision signal for NeRF-style gaze prediction. 
\item We propose a framework that reconstructs gaze patterns from decoupled observer and rendering camera perspectives, explicitly modeling gaze occlusion through NeRF-based depth testing.
\item We evaluate NeRGs in a real-world convenience store case study, where head pose tracking data serves as a proxy for gaze, learning a geometry-aware 3D gaze representation that supports novel view synthesis and interactive visualization of 3D gaze patterns.
\end{itemize}

\section{Related Work}

\subsectionnonum{Visual Salience.}
Visual salience is commonly represented by 2D salience maps that highlight image regions likely to attract attention \cite{buswellAttention1935}. Early computational models relied on low-level features such as intensity, color, and edge orientation \cite{itti1998model, koch1987shifts}, using various strategies for feature extraction and refinement \cite{harel2006graph, zhang2008sun, judd2009learning, zhang2013saliency, adeli2017model}. More recently, deep learning has become central to salience prediction \cite{kummerersaliencyBenchmarking, vig2014large, deepFeaturesSalience2015, kummerer2017understanding, kruthiventi2017deepfix, TranSalNet2022}, with models trained on eye-tracking data from free-viewing and task-specific settings \cite{salicon2015, mitSaliencyBenchmark}. A notable example is the DeepGaze family of models \cite{kummerer2014deep, kummerer2016deepgaze, kummerer2022deepgaze}, which leverage deep features from pre-trained CNNs to predict salience or eye fixation positions in 2D images. 

\subsectionnonum{Salience in 3D.} 
Eye-tracking studies \cite{krafka2016eye, macinnes2018wearable, ye2012detecting} and gaze-target prediction from third-person videos \cite{chong2018connecting, chong2020detecting, kellnhofer2019gaze360, nonaka2022dynamic} have largely focused on 2D image analysis, which differs fundamentally from salience in 3D environments. Complex 3D salience cannot be fully captured by 2D methods \cite{wang2018slam, yang2021visual, elmadjian20183d, song20233d}. 
Some work has projected 2D salience onto 3D scenes \cite{maurus2014realistic, attention3Dshapes}, while others have modeled mesh salience to characterize attention over 3D objects \cite{lee2005mesh, song2019, mesh3DSaliency2024}.
Studies in virtual reality (VR) and virtual 3D environments further show that fixations often occur at low head velocities and largely align with head pose, suggesting that 3D salience can be inferred from head orientation even without explicit gaze tracking \cite{gazeAllocation2011, johnson2013predicting}.

\subsectionnonum{Egocentric Gaze Estimation.}
Egocentric gaze estimation methods model attention from the viewpoint of an observer \cite{zhang2017deep, lai2023eye, tavakoli2019digging}, typically using first-person video to analyze and predict gaze patterns \cite{huang2018predicting, huang2020ego, naas2020functional, thakur2021predicting}. These approaches remain inherently 2D, predicting in image space rather than within a 3D scene context. An additional challenge is high frame-to-frame variability, which lowers temporal consistency and requires advanced filtering or task-driven modeling to separate true attention shifts from noise or unrelated movements.

\subsectionnonum{Neural Radiance Fields.}
Neural Radiance Fields (NeRFs) \cite{mildenhall2021nerf} represent 3D scenes as continuous volumetric functions learned from 2D images with known camera parameters. Given a position $(x,y,z)$ and viewing direction $(\theta,\phi)$, NeRFs evaluate a volume density $\sigma$ and a view-dependent RGB color, which are integrated along camera rays via volume rendering to produce pixel values. Models are trained by minimizing the perceptual difference between rendered and ground-truth views. Recent advances have improved NeRF rendering speed and quality \cite{barron2021mip, barron2022mip360, barron2023zip, mueller2022instantNgp}, solidifying their role as a powerful scene representation.

\subsectionnonum{NeRFs and 3D Semantics.}
Beyond novel view synthesis, NeRF-based methods have been extended to semantic reasoning in 3D, associating geometry with object categories or part labels \cite{semanticNerf2021, segmentAnythingNerfs2023, panopticNerf2022}. Notably, Semantic-NeRF \cite{semanticNerf2021} integrates semantic features along each ray to jointly predict class labels and RGB color but requires retraining each NeRF to incorporate semantics. More recent work explores generalizable methods that operate on pretrained NeRFs in a near zero-shot setting \cite{semanticRay2023, nesf2022, instanceNerf2023}. However, these approaches primarily focus on static object categories with local labels invariant to view direction and observer position. In contrast, human visual attention depends on the visual field of the observer, motivating models that couple view-dependent perception with 3D geometry.

\section{Preliminaries}

\begin{figure}[t!]
\centering
    \includegraphics[width=0.495\linewidth]
    {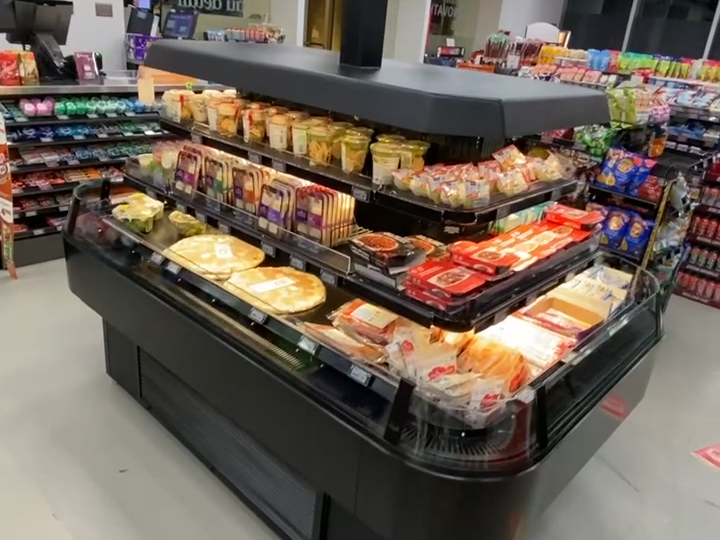}
    \includegraphics[width=0.495\linewidth]
    {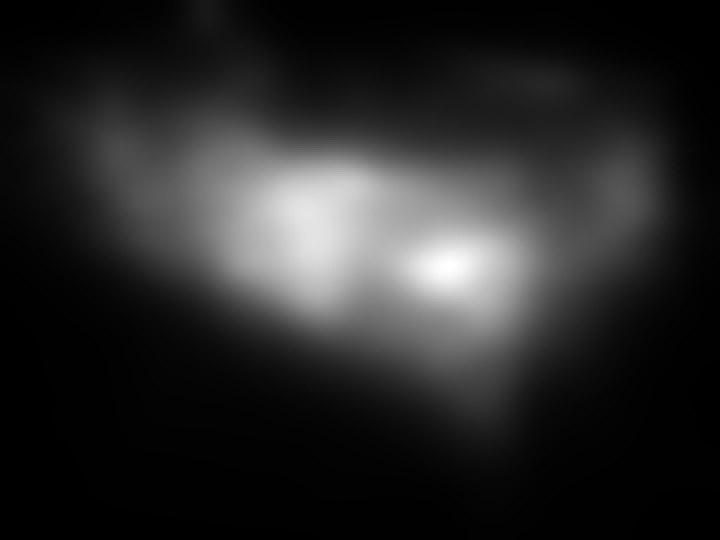}
\caption{Visualization of 2D salience for a 3D scene. (Left) Scene rendered with a pre-trained NeRF. (Right) Gaze probability density predicted by DeepGaze IIe \cite{linardos2021deepgaze}, a salience model that infers attention from 2D pixels without accounting for 3D geometry.}
\label{fig:deepgaze}
\end{figure}

\begin{figure*}[t!]
\centering
\includegraphics[width=0.8\textwidth, trim={0.5cm 0.25cm 0.25cm 0.5cm}, clip]{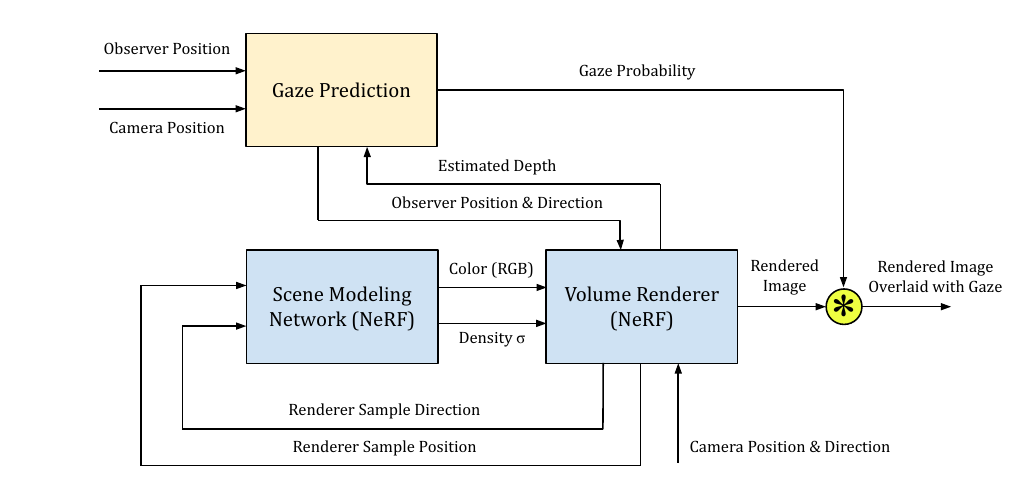}
\caption{Overall system diagram of the Neural Radiance and Gaze Field (NeRG). We evaluate a NeRF with volume rendering and estimate gaze for the visible surfaces. Gaze is predicted from the observer's perspective, which can be decoupled from the rendering camera.}
\label{fig:nerg-system}
\end{figure*}

A straightforward approach to incorporating gaze into NeRFs is to apply a 2D salience model, such as DeepGaze \cite{kummerer2016deepgaze}, to scene renders (see \autoref{fig:deepgaze}). However, 2D salience methods require 2D image inputs and do not consider the underlying 3D geometry, which is critical for accurately modeling gaze in 3D \cite{mesh3DSaliency2024}. These methods also incur substantial computational cost, as each frame must be processed by a salience network, and may yield inconsistent predictions on NeRF-rendered images due to visual artifacts (e.g., floaters) or non-standard camera orientations. Mapping 2D salience to 3D via depth projection further constrains the observer to the rendering camera’s position, preventing decoupling of the two perspectives.

Instead, we predict 3D gaze within the NeRF-based scene representation, directly leveraging it as a geometric prior. Our objective is to evaluate gaze in the 3D scene with potentially decoupled observer and rendering perspectives, enabling flexible visualization from arbitrary viewpoints. Because regions visible to the rendering camera may be occluded from a decoupled observer’s view, modeling gaze occlusion is a key component of our approach.

\section{Neural Radiance and Gaze Fields} 
\label{sec:nerg}

\begin{figure}[t]
\centering
\includegraphics[width=\linewidth, trim={6.25cm 3.5cm 6.75cm 2cm}, clip]{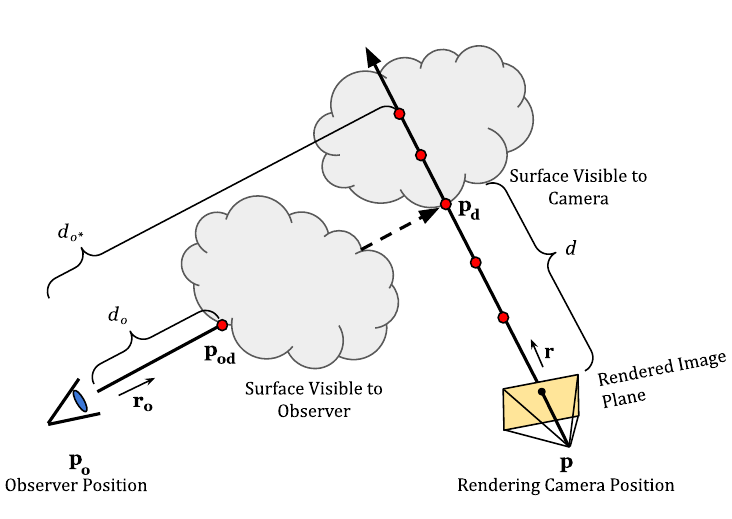}
\caption{Gaze prediction in NeRG. Volume rendering of the NeRF produces a depth estimate along the traced ray. Gaze is evaluated between the position of the visible surface and the position of the observer. If the observer is decoupled from the rendering camera, gaze occlusion can be modeled by additionally evaluating depth from the observer's perspective to correct the position of the surface visible to the observer.}
\label{fig:nerf-gaze}
\end{figure}

We present Neural Radiance and Gaze Fields (NeRGs), a novel method for representing and visualizing gaze in 3D environments. Our approach extends NeRFs with a gaze prediction module (see \autoref{fig:nerg-system}), constraining gaze prediction to the 3D scene geometry encoded by a pre-trained NeRF. As illustrated in \autoref{fig:nerf-gaze}, we first evaluate the NeRF from position $\mathbf{p}$ along ray direction $\mathbf{r}$ to obtain its color via volume rendering. The depth $d$ along the traced ray defines the surface at position $\mathbf{p_{d}} = \mathbf{p} + d\mathbf{r}$, which is visible from the rendering perspective. While NeRFs do not explicitly encode depth, a coarse estimate can be recovered by accumulating density during volume rendering \cite{mildenhall2021nerf}. We then compute gaze by evaluating a lightweight network for a gaze ray from the observer position $\mathbf{p_{o}}$ toward $\mathbf{p_{d}}$.
Unlike prior methods, our system decouples the observer position from the rendering camera, enabling more flexible visualizations and finer user control. 
When the observer is decoupled, NeRG handles gaze occlusion with an additional depth test from the observer's perspective: a discrepancy between the expected and evaluated depth values implies occlusion along the gaze ray. 

\subsection{Modeling 3D Gaze}
\label{sec:modeling_gaze}

Our system models the gaze signal between an observer and a visible surface in the 3D scene. In NeRG, surfaces \textit{emit} a signal that encodes their geometric or semantic features, and observers \textit{capture} this signal through gaze allocation. The emitted signal serves to constrain gaze prediction by enforcing consistency with the 3D scene geometry and surface visibility, whereas the capture term conditions gaze on the observer’s position and visual field, introducing a perceptual, view-dependent component.

Prior NeRF-based methods \cite{semanticNerf2021, semanticRay2023} focus on 3D segmentation or view-invariant object labels, whereas NeRG models surface-level signals as perceived from a specific observer viewpoint. Moreover, NeRG allows the observer to be decoupled from the rendering perspective, modeling gaze occlusion via a visibility test from the observer’s position using NeRF-based depth. Analogous to shadows from light occlusion, the emitted signal can be blocked from an observer, preventing gaze allocation to surfaces visible only to the rendering camera. 

\subsection{Gaze Prediction Module}
\label{sec:gaze_prediction}

Following NeRF, we approximate a continuous spatial function parameterized by an MLP neural network \mbox{$F_\Theta : (\mathbf{p}, \mathbf{r}) \rightarrow g$} for input position \mbox{$\mathbf{p}=(x,y,z)$} and ray direction \mbox{$\mathbf{r}=(\theta,\phi)$}. Instead of density and color (as in NeRF), our model encodes a directional gaze signal $g$. We further define two identical networks, $F_e$ and $F_c$, to model the emitted and captured signals related to gaze.
The network $F_e$ encodes the signal emitted from position $\mathbf{p_{od}}$ (surface visible to observer). The network $F_c$ models the portion of emitted signal captured by an observer at position $\mathbf{p_{o}}$.
Gaze prediction is then defined as follows: 
\begin{equation}
\label{eqn:gaze_prediction}
    G(\mathbf{p_{od}}, \mathbf{p_o}) = F_e(\mathbf{p_{od}}, -\mathbf{r_o}) \times \text{sigmoid} ( F_c(\mathbf{p_o}, \mathbf{r_o}) ),
\end{equation}
where $\mathbf{r_o}$ describes the direction from $\mathbf{p_{o}}$ to $\mathbf{p_{od}}$. 

We apply a sigmoid activation function to the output of $F_c$ to represent the probability that an emitted signal is perceived by the observer. 
The capture component is necessary to model spatial variations in gaze patterns along a ray from a given surface. The emit component alone cannot achieve this, as the emitted signal remains constant along a given direction, similar to radiance or semantic labels.
In practice, networks $F_e$ and $F_c$ share most weights, embedding input position and direction into a common latent space (similar to geometric features in NeRF), but use separate final fully connected layers to produce their respective outputs. 

To reduce computational overhead and keep the model lightweight, NeRG performs gaze prediction without volume rendering: we do not accumulate gaze values along a ray, as is done for density and color in NeRF or semantic features in Semantic-NeRF \cite{semanticNerf2021}. Although we tested variants with accumulated semantic features from pre-trained and fine-tuned NeRF geometry, depth-based feature computation was sufficient. Moreover, depth can be computed efficiently by evaluating only the density component of NeRF. Lastly, when the observer and rendering perspectives are decoupled, NeRG also requires tracing depth from the observer's perspective. 

\subsection{Surface Visibility and Gaze Occlusion}
\label{sec:shadow}

When the observer and rendering camera perspectives are aligned, depth is equivalent from the two perspectives, and gaze occlusion does not need to be handled. When the perspectives are decoupled, similarly to how objects cast shadows to occlude light, the emitted signal from a surface can be occluded from an observer. 
We test for gaze occlusion by evaluating depth from the observer perspective and comparing it to the expected depth. 
We evaluate depth with an efficient volume rendering pass by the NeRF (only accumulating volumetric density values $\sigma$). Although NeRF-estimated depth may be inconsistent, even in regions where it is expected to be uniform, we find it sufficient for modeling visibility in the encoded 3D environment. 

As illustrated in \autoref{fig:nerf-gaze}, depth $d$ is measured from the rendering camera position, depth $d_o$ is the evaluated depth from the observer perspective, and $d_{o*}$ is the expected depth given by $\|\mathbf{p_{o}}-\mathbf{p_{d}}\|$.
When $d_o < d_{o*}$, the surface visible to the camera is occluded from the perspective of the observer, implying gaze occlusion. This comparison acts as a binary indicator of visibility from the observer. For a smoother transition, we define a gaze shadowing factor $s$ with a fall-off parameter $d_f$, namely $s=(d_{o*} - d_o) / d_f$. A larger fall-off value produces a gaze penumbra effect which is visually preferable to the sharp and high variance shadowing resulting from a boolean comparison. We finalize by clipping $0 < s < 1$ and multiplying gaze probability by the gaze shadowing factor.

\subsection{Training and Loss Functions}
\label{sec:loss_functions}

NeRG employs a fixed pretrained NeRF and augments it with a lightweight network for gaze prediction. To train this network, we optimize standard loss functions used in saliency prediction \cite{TranSalNet2022, kummerersaliencyBenchmarking, saliencyEvalMetrics}, given the ground-truth and predicted gaze probability density values $\mathbf{y}$ and $\mathbf{\hat{y}}$, respectively. Specifically, we employ the Kullback-Leibler divergence (KLD), the linear correlation coefficient (CC), and the mean absolute error (MAE), defined as:
\begin{equation}
    L_{\text{CC}}(\mathbf{y}, \mathbf{\hat y}) =  \frac{\text{cov}(\mathbf{y}, \mathbf{\hat y})}{\sigma(\mathbf{y}) \sigma(\mathbf{\hat y)}},
\end{equation}
where $\text{cov}(\cdot)$ is covariance and $\sigma(\cdot)$ is standard deviation;
\begin{equation}
    L_{\text{KLD}}(\mathbf{y}, \mathbf{\hat{y}}) = \sum_{i=1}^{N} \mathbf{y_i} \text{log}\left( \frac{\mathbf{y_i}}{\mathbf{\hat{y}_i}} \right),
\end{equation}
where $N$ represents the number of samples;
\begin{equation}
    L_{MAE}(\mathbf{y}, \mathbf{\hat{y}}) = \frac{1}{N} \sum_{i=1}^{N} | \mathbf{y_i} - \mathbf{\hat{y}_i} |.
\end{equation}
As in \cite{TranSalNet2022}, we optimize the sum of individual loss terms.

\section{Experiments}

\subsection{Obtaining Training Data}

\begin{figure}[t]
  \centering
  \includegraphics[width=\linewidth, trim={2cm 2cm 1.5cm 3cm}, clip]{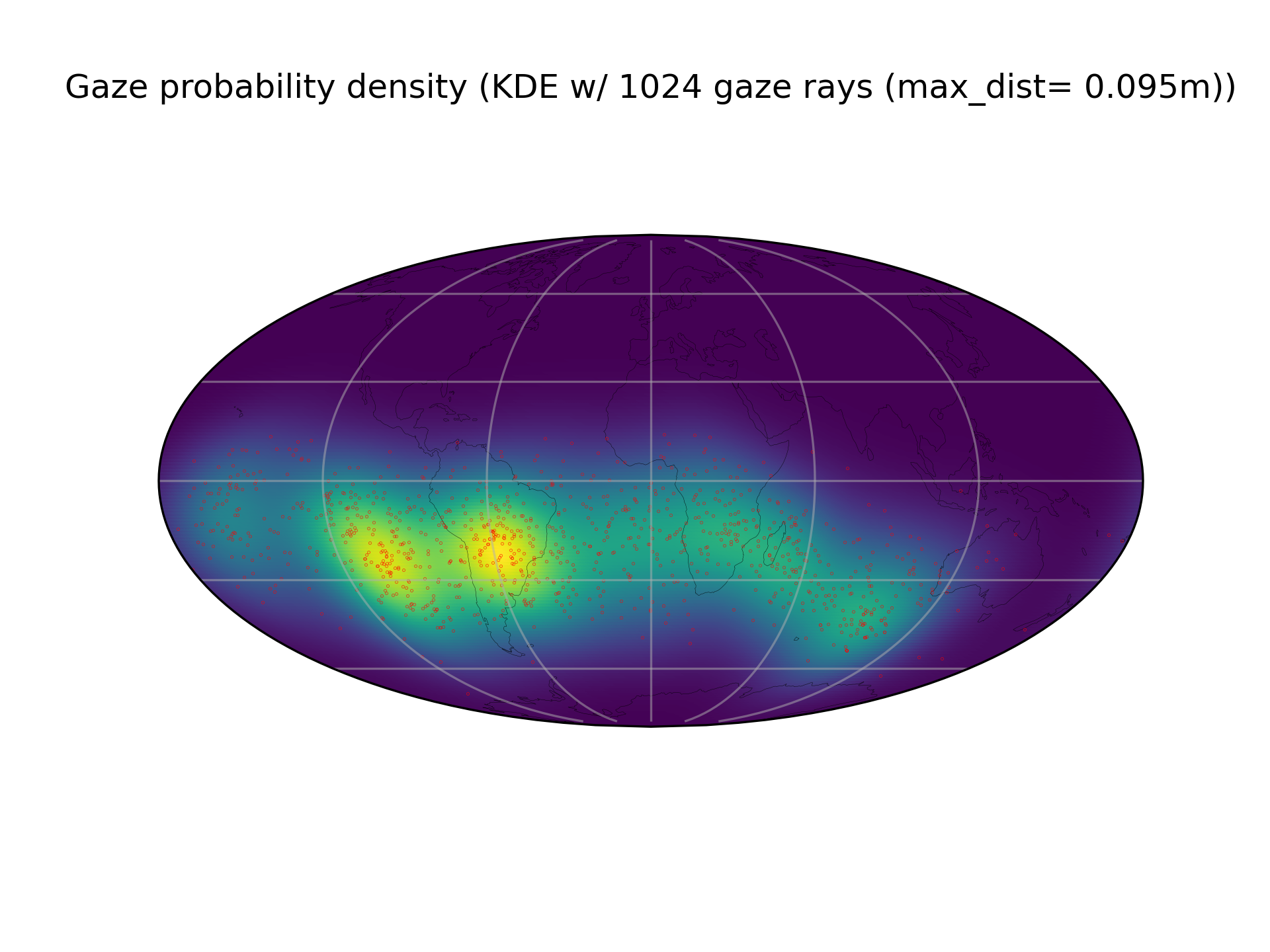}
  \caption{
  Egocentric gaze probability density for a randomly selected gaze probe (data from \cite{Wang2024eyeTracking}), computed from the raw head pose rays (red dots) and visualized as a spherical heatmap.
  }
\label{fig:gaze_probe}
\end{figure}

Most gaze and salience datasets do not provide the multi-view coverage needed for NeRF-style reconstruction, making it challenging to train and evaluate NeRG under the experimental setups commonly used in existing studies. As a practical alternative, we leverage the more widely available skeleton-tracking data collected in known 3D environments. In retail settings, for instance, in-store tracking systems such as C2RO’s ENTERA \cite{C2RO} capture and analyze customer movement and behavioral patterns, providing large-scale head pose data. Following prior work \cite{gazeAllocation2011, johnson2013predicting}, we adopt the simplifying assumption that head pose approximates gaze direction. To train NeRG, we construct a probabilistic representation by aggregating noisy individual head pose rays into \emph{gaze probes} that model local egocentric gaze probability densities. While NeRG could also be trained using 2D views and gaze predictions from 2D salience models, such methods do not adequately capture gaze in complex 3D scenes and task-dependent viewing. 

\begin{figure}[t]
  \centering
  \includegraphics[width=\linewidth]{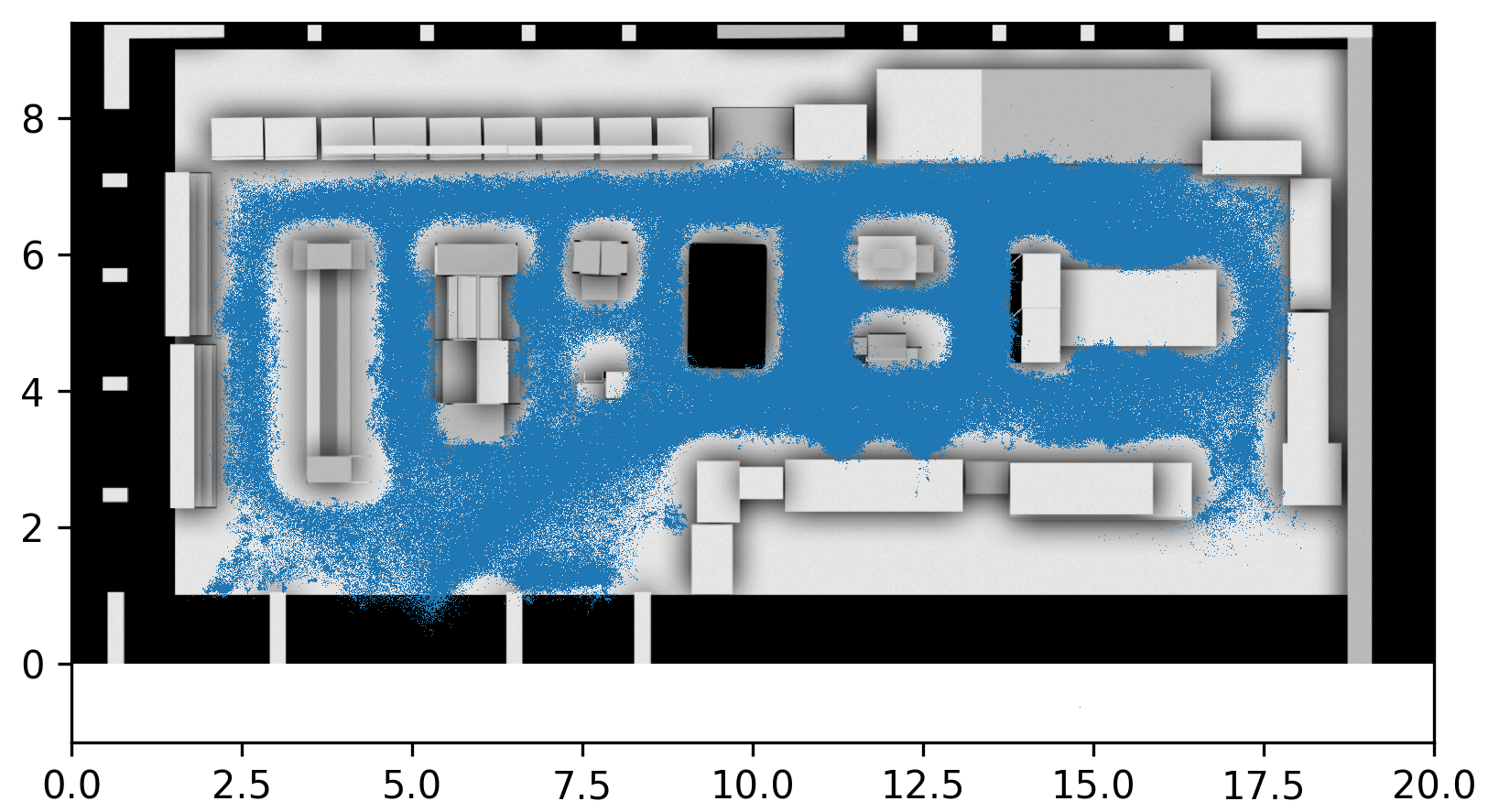}
  \caption{Layout and dimensions (in meters) of the convenience store corresponding to the pose-tracking data from \cite{Wang2024eyeTracking}. The blue point cloud shows the positions of individual gaze rays.
  }
\label{fig:store_layout}
\end{figure}

\subsectionnonum{Dataset.} 
We use a large-scale skeleton-tracking dataset collected in a convenience store and presented by Wang et al. \cite{Wang2024eyeTracking}. The dataset consists of customer trajectories recorded as skeleton pose sequences, with 22 tracked points on the human body, termed \textit{keypoints}. To estimate head pose, we extract keypoints for the nose and the left and right ears, then compute a directional vector to approximate gaze direction. We use a one-day subset of the tracking data (approximately 2.6M individual gaze rays) spatially located in the central area of the store which matches the views used for NeRF training, as illustrated in \autoref{fig:store_layout}. In our study, we first train a high-quality NeRF model to represent the 3D environment and then extend it with gaze prediction using our proposed NeRG. 

\subsectionnonum{Gaze probes.}
\label{sec:gaze-data}
To obtain a representation suitable for training NeRG, we group individual head pose rays into gaze probes, each aggregating rays within a physical radius $R$ of a given position in the 3D environment. 
For each probe, we apply kernel density estimation (KDE) to its rays to compute the gaze probability density over viewing directions on the unit sphere. For the kernel, we adopt the von Mises-Fisher (vMF) spherical distribution \cite{vonmissesfisher}, which is well suited for directional data. A gaze probe models a valid local 360$^{\circ}$ egocentric gaze probability density function that integrates to one, with an example shown in \autoref{fig:gaze_probe}. 
Compared to individual head pose rays, our gaze probes provide a more robust and spatially stable supervision signal by aggregating noisy measurements into properly normalized local probability distributions. 

\subsectionnonum{Training data.} We begin by generating gaze probes, placing them either uniformly on a 3D grid within the scene or by randomly sampling positions.
For each gaze probe, training samples are obtained by uniformly sampling directions on the sphere and evaluating the corresponding gaze probability density using KDE. 
NeRG is then trained on a set of gaze ray samples $s_i:\{x,y,z,\theta,\phi,g\}$, where $(x,y,z)$ is given by the corresponding gaze probe position, $(\theta,\phi)$ is the gaze ray direction, and $g$ is the ground-truth gaze probability density. This formulation follows the setup used to train NeRFs, except that NeRGs predict gaze probability density instead of color values. 
Note that gaze occlusion does not occur during training, since the rendering and observer perspectives are always aligned in this setup. 

\subsectionnonum{Aligning NeRF and gaze data.}
Gaze prediction in NeRF-space requires knowing the mapping between the NeRF coordinate system and the world coordinate system in which the pose-tracking data is collected.
To obtain this mapping, we leverage a 3D digital twin (3D mesh) of the convenience store, viewable in 3D modeling software such as Blender\textsuperscript{\textregistered} \cite{blender}. We first extract a coarse 3D mesh from the trained NeRF using the marching cubes algorithm \cite{mueller2022instantNgp}, expressed in NeRF coordinates. This mesh is manually aligned with the digital twin in Blender to recover the world transformation matrix $M_{w}$, representing the mapping from NeRF to gaze coordinates. Determining this mapping automatically is challenging, and we leave this to future work.

\subsection{Implementation}

We use Pytorch \cite{pytorch} and build on top of a public repository of NeRF\footnote{\url{https://github.com/ashawkey/torch-ngp}} based on instant neural primitives \cite{mueller2022instantNgp}. In our experiments, we train NeRG on a single NVIDIA GeForce RTX 3090 GPU using the Adam optimizer \cite{adam} (hyperparameters as recommended in \cite{mueller2022instantNgp}) for 50 epochs on 4,096 gaze probes, with 1,024 uniform sphere samples per probe per epoch. A separate set of 512 gaze probes is held out for testing, similar to NeRF evaluations where training and testing views differ in location but share the same scene. To construct gaze probes, we collect head pose rays within \mbox{$R=0.1$} meters of probe center and cap the maximum number of gaze rays to 1024 per probe when computing KDE. For efficiency, we precompute and reuse two fixed sets of gaze probes (training and testing), since active indexing into a large ray dataset is computationally expensive. 
We release our code at {\href{https://github.com/ch-andrei/nerg-neural-gaze}{https://github.com/ch-andrei/nerg-neural-gaze}}.


\subsection{Evaluations and Ablations}

\begin{figure}[t]
  \centering
  \includegraphics[width=\linewidth, trim={0.25cm 0.2cm 0.25cm 0cm}, clip]{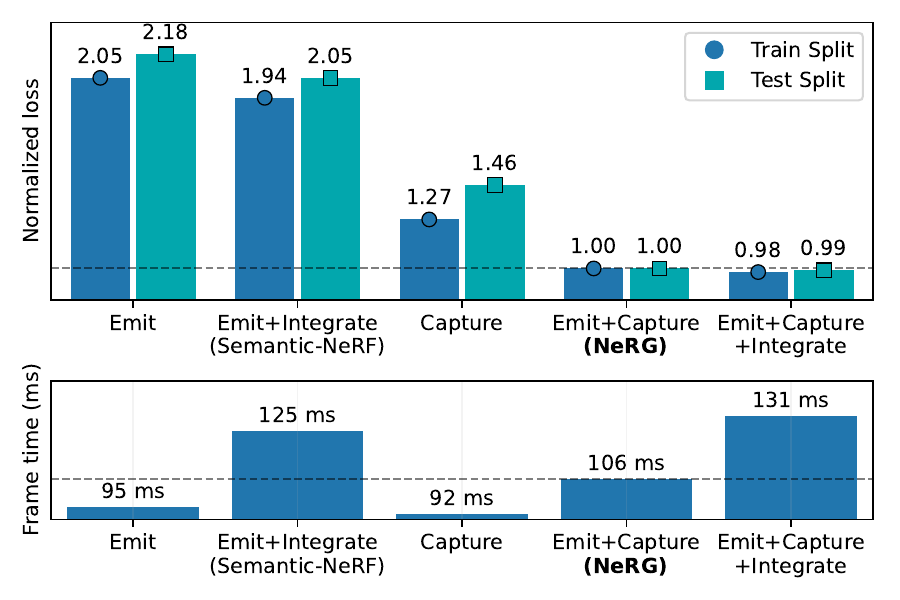}
  \caption{Loss metrics and 720p inference frame time for NeRG variants. Loss values are normalized relative to the \textit{Emit+Capture} variant, which achieves the best performance trade-off, conditioning gaze prediction on both the observer and the visible surface. Integrating semantic features via volume rendering (similar to Semantic-NeRF \cite{semanticNerf2021}) produces negligible accuracy gains while increasing frame time. All models use comparable configurations.
  }
\label{fig:train_test_loss}
\end{figure}

We evaluate NeRG on a held-out set of gaze probes, comparing ground-truth and predicted gaze patterns using the loss functions defined in \autoref{sec:loss_functions} as metrics. To assess computational cost, we record the average frame time (FT) to render a 720p scene, including both NeRF and gaze prediction forward passes, measured over 32 randomly placed cameras with shared positions across models. The results of our tests are summarized in \autoref{fig:train_test_loss}. 

Our evaluations highlight the importance of the emit-capture formulation defined by \autoref{eqn:gaze_prediction}, which significantly improves the consistency of reconstructed gaze patterns, particularly on the held-out test set. 
The \textit{Emit} variant has the lowest accuracy because it does not model the observer, whereas the \textit{Capture} variant is slightly better but overfits the training views due to the absence of geometric priors. We achieve the best performance with the \textit{Emit+Capture} variant, which conditions gaze prediction on both the visible surface and the observer position.

We also find that integrating semantic features from a pre-trained or fine-tuned NeRF via volume rendering, following the Semantic-NeRF formulation \cite{semanticNerf2021}, does not produce substantial improvements over NeRG without volume rendering.
Semantic-NeRF is conceptually similar to the \textit{Emit} variant of NeRG but includes integration of multiple emitted features through volume rendering along a ray. In contrast, NeRG computes a single snapshot of geometric features at the predicted depth, which we find sufficient. 

Finally, for network configuration in NeRG, we observe that the standard NeRF architecture and its default hyperparameters produce the best performance. Consequently, the gaze prediction module in NeRG uses the same number of layers, hidden dimensions, activation functions, and other related settings as the NeRF network. Although we experimented with more complex designs, the baseline NeRF configuration achieved the strongest results. Since NeRG reuses the NeRF architecture for the networks $F_e$ and $F_c$ (see \autoref{sec:gaze_prediction}), we also tested initializing the geometric feature modules with weights from the pre-trained NeRF model, both frozen and fine-tuned, but observed no clear improvement over training from scratch. 

\begin{figure*}[t]
     \centering
     \begin{subfigure}[b]{0.33\textwidth}
         \centering
         \includegraphics[width=\textwidth, trim={1cm 1cm 1cm 1cm}, clip]
         {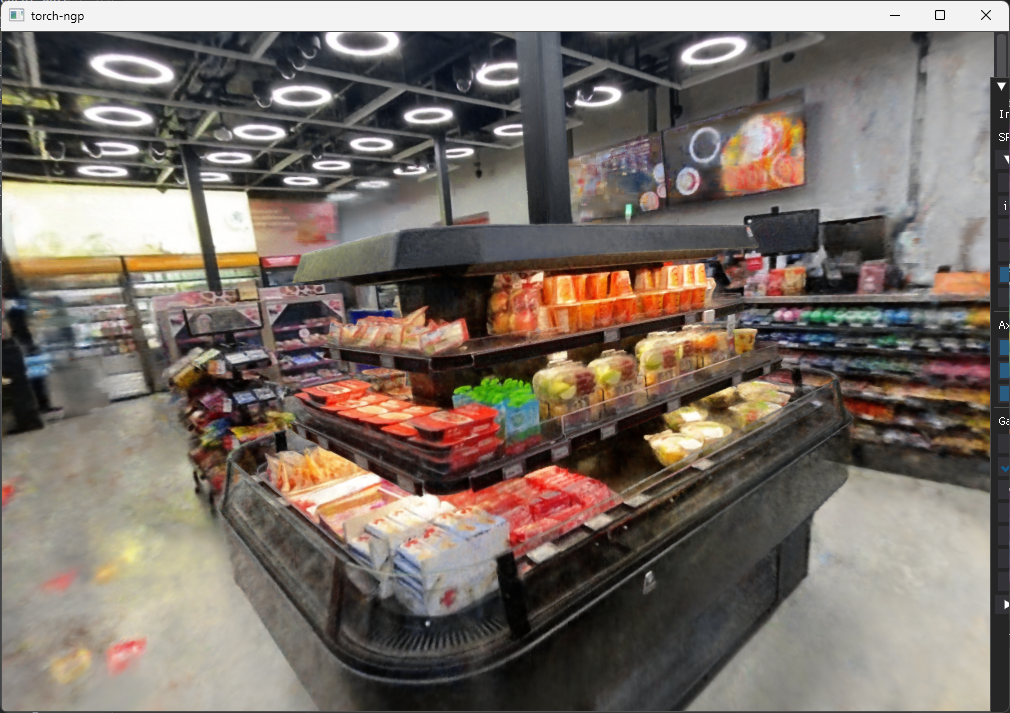}
         \caption{}
     \end{subfigure}
     \hfill
     \begin{subfigure}[b]{0.33\textwidth}
         \centering
         \includegraphics[width=\textwidth, trim={1cm 1cm 1cm 1cm}, clip]
         {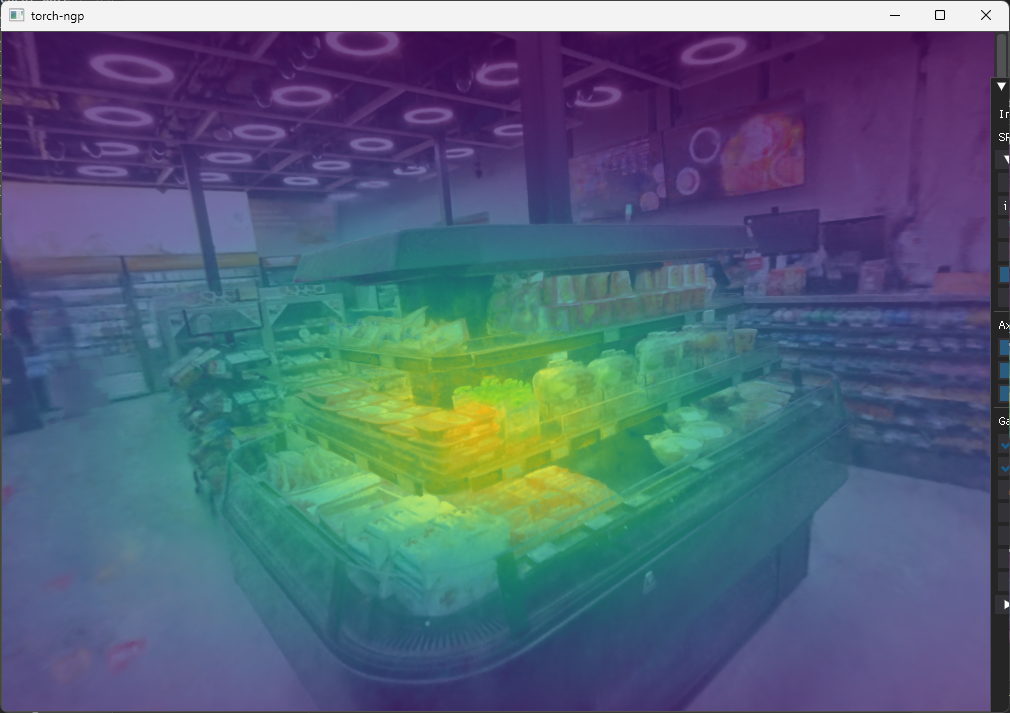}
         \caption{}
     \end{subfigure}
     \hfill
     \begin{subfigure}[b]{0.33\textwidth}
         \centering
         \includegraphics[width=\textwidth, trim={1cm 1cm 1cm 1cm}, clip]
         {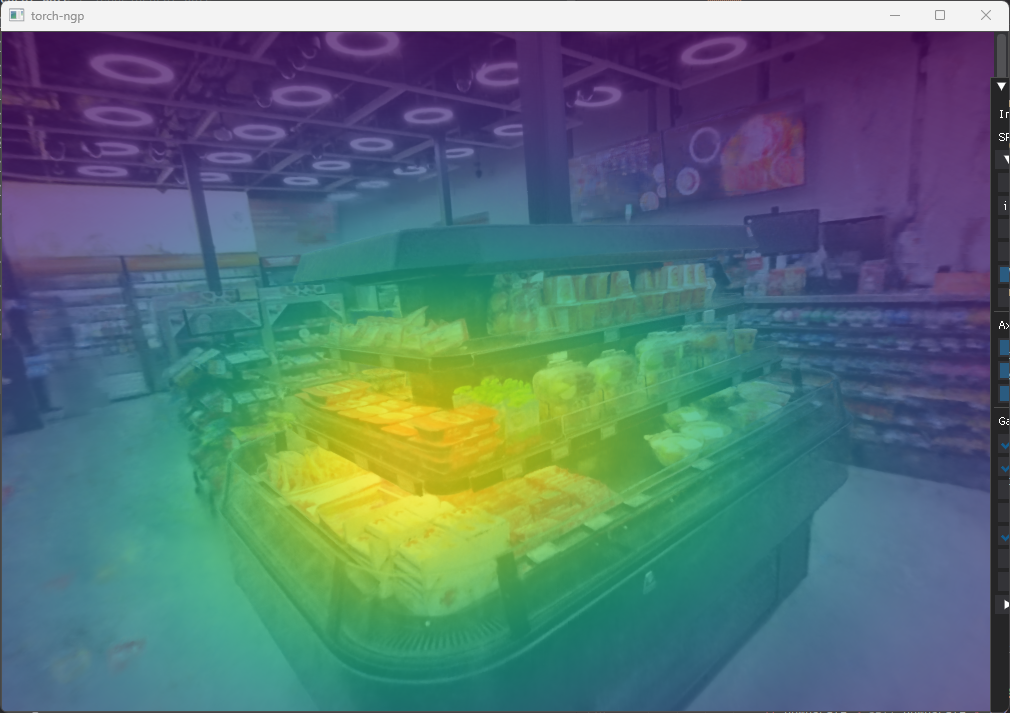}
         \caption{}
     \end{subfigure}
     \begin{subfigure}[b]{0.33\textwidth}
         \centering
         \includegraphics[width=\textwidth, trim={1cm 1cm 1cm 1cm}, clip]
         {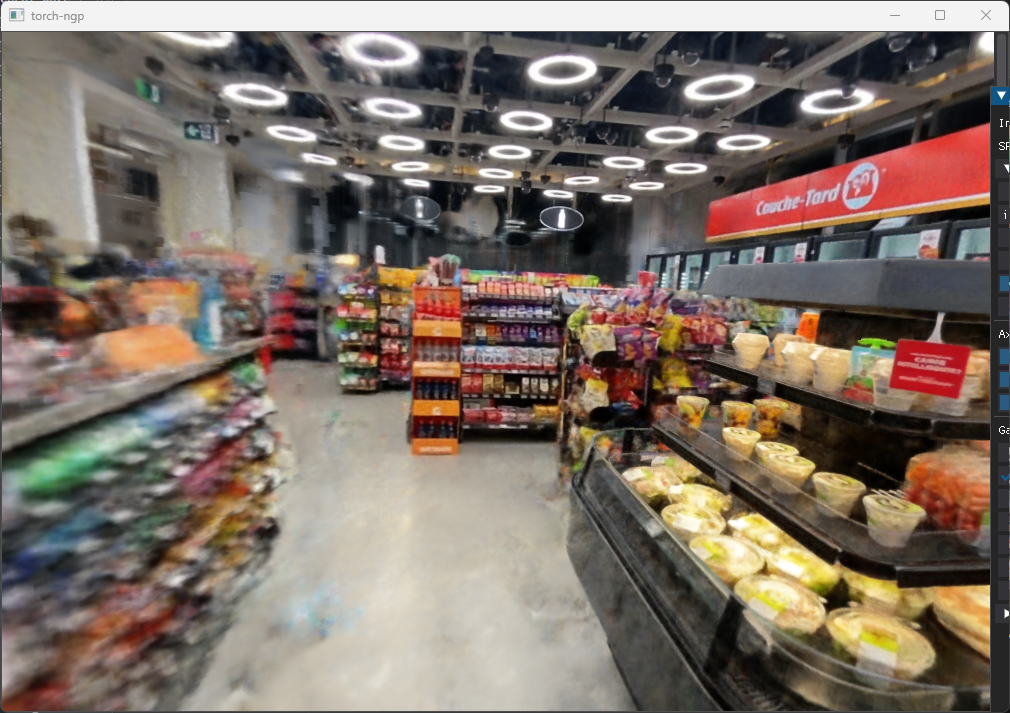}
         \caption{}
     \end{subfigure}
     \hfill
     \begin{subfigure}[b]{0.33\textwidth}
         \centering
         \includegraphics[width=\textwidth, trim={1cm 1cm 1cm 1cm}, clip]
         {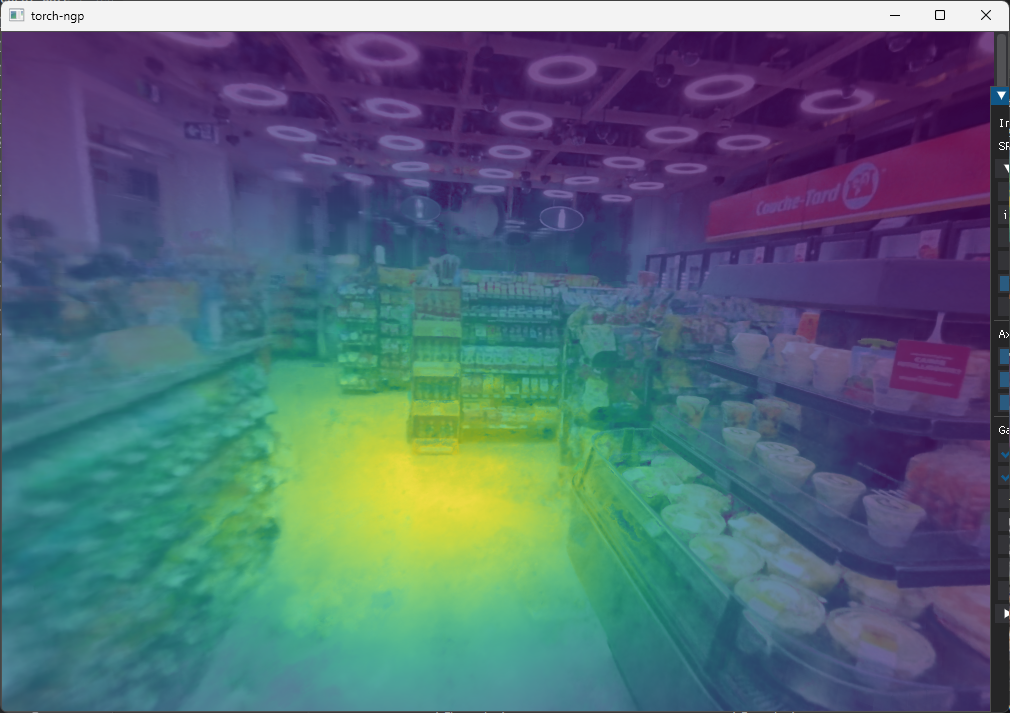}
         \caption{}
     \end{subfigure}
     \hfill
     \begin{subfigure}[b]{0.33\textwidth}
         \centering
         \includegraphics[width=\textwidth, trim={1cm 1cm 1cm 1cm}, clip]
         {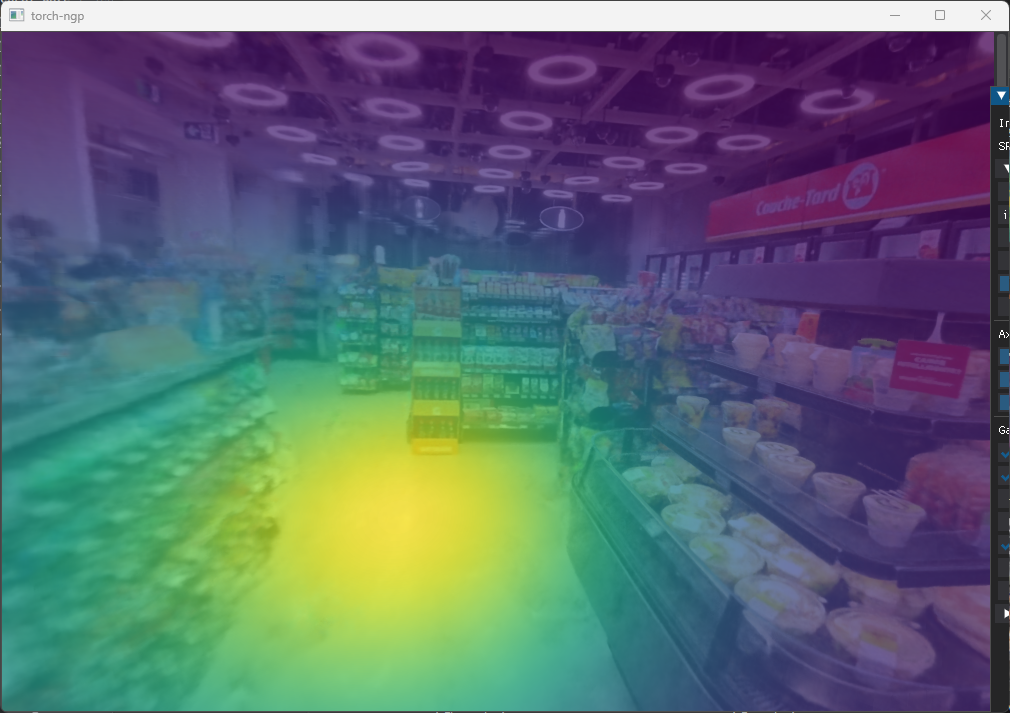}
         \caption{}
     \end{subfigure}
        \vspace{-0.75cm}
        \caption{Gaze visualization for NeRG and gaze probes: (a) and (d) NeRF-rendered views, (b) and (e) 3D gaze predictions by NeRG, (c) and (f) ground-truth gaze density from gaze probes.
        Gaze probes aggregate local gaze ray data and use kernel density estimation (KDE) to compute gaze probabilities, which is computationally expensive for large ray datasets (5+ seconds per 720p frame for data from \cite{Wang2024eyeTracking}). In contrast, NeRG recovers spatially consistent 3D gaze patterns and renders them interactively (under 100 ms per frame).
        } 
        \label{fig:gaze}
\end{figure*}

\begin{figure*}[!t]
     \centering
     \begin{subfigure}[b]{0.245\textwidth}
         \centering
         \includegraphics[width=\textwidth, trim={5cm 0.1cm 1cm 3cm}, clip]
         {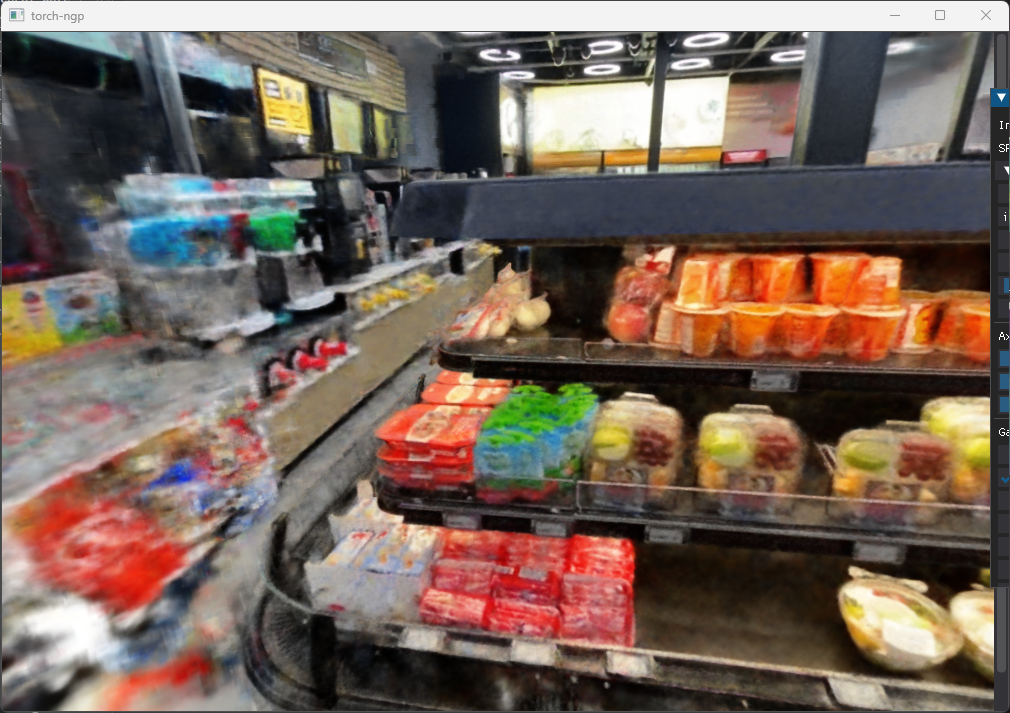}
         \caption{}
     \end{subfigure}
     \hfill
     \begin{subfigure}[b]{0.245\textwidth}
         \centering
         \includegraphics[width=\textwidth, trim={5cm 0.1cm 1cm 3cm}, clip]
         {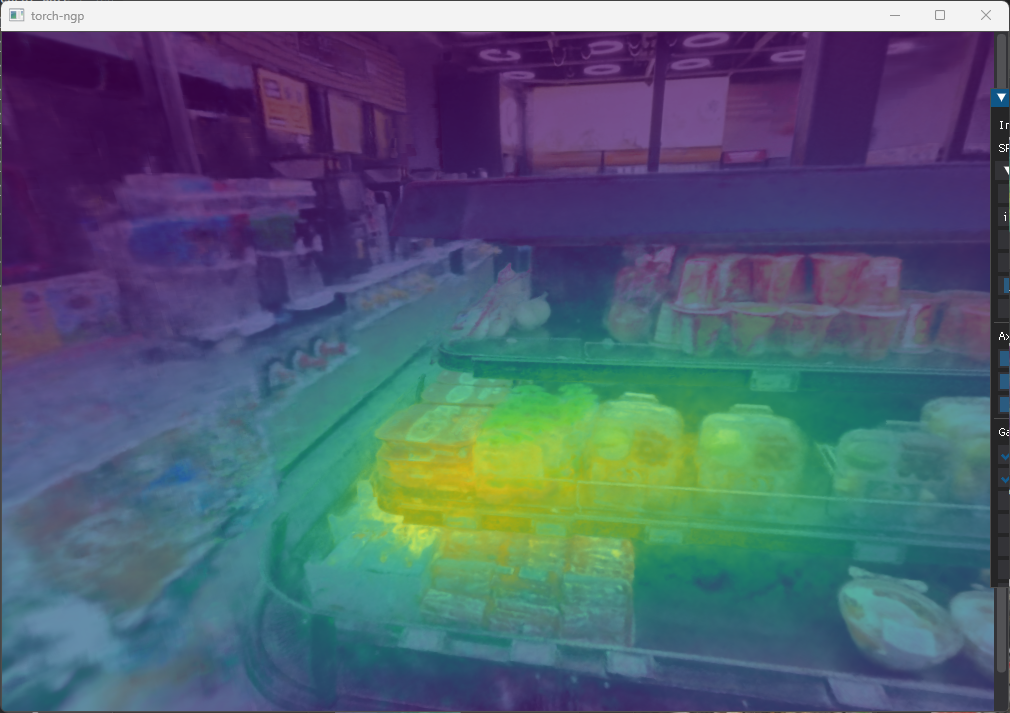}
         \caption{}
     \end{subfigure}
     \hfill
     \begin{subfigure}[b]{0.245\textwidth}
         \centering
         \includegraphics[width=\textwidth, trim={5cm 0.1cm 1cm 3cm}, clip]
         {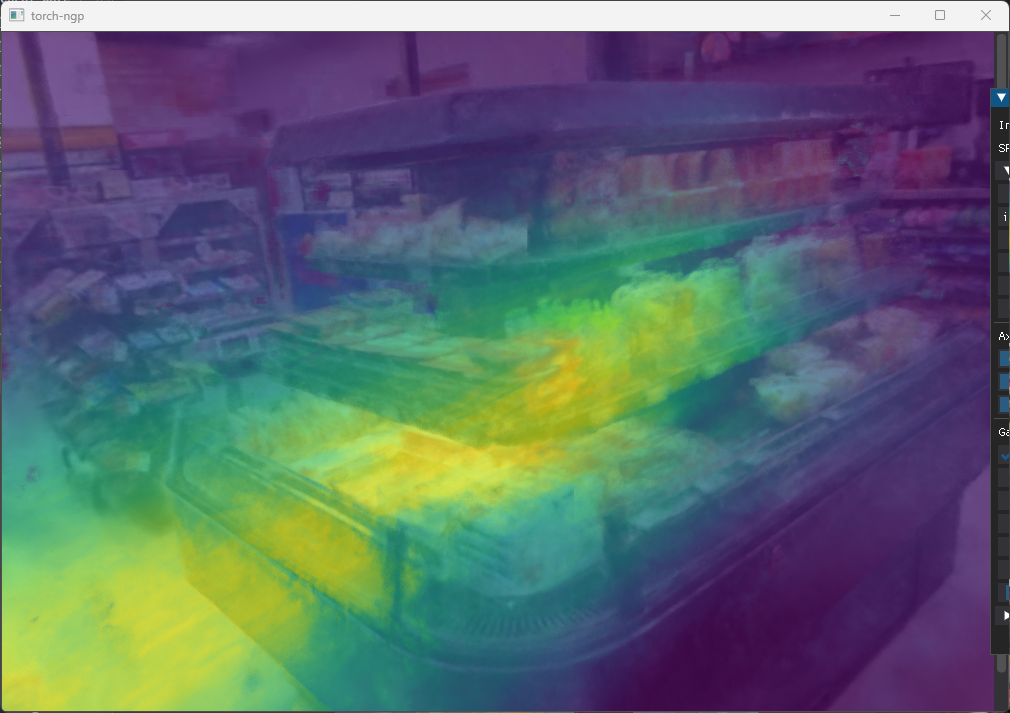}
         \caption{}
     \end{subfigure}
     \begin{subfigure}[b]{0.245\textwidth}
         \centering
         \includegraphics[width=\textwidth, trim={5cm 0.1cm 1cm 3cm}, clip]
         {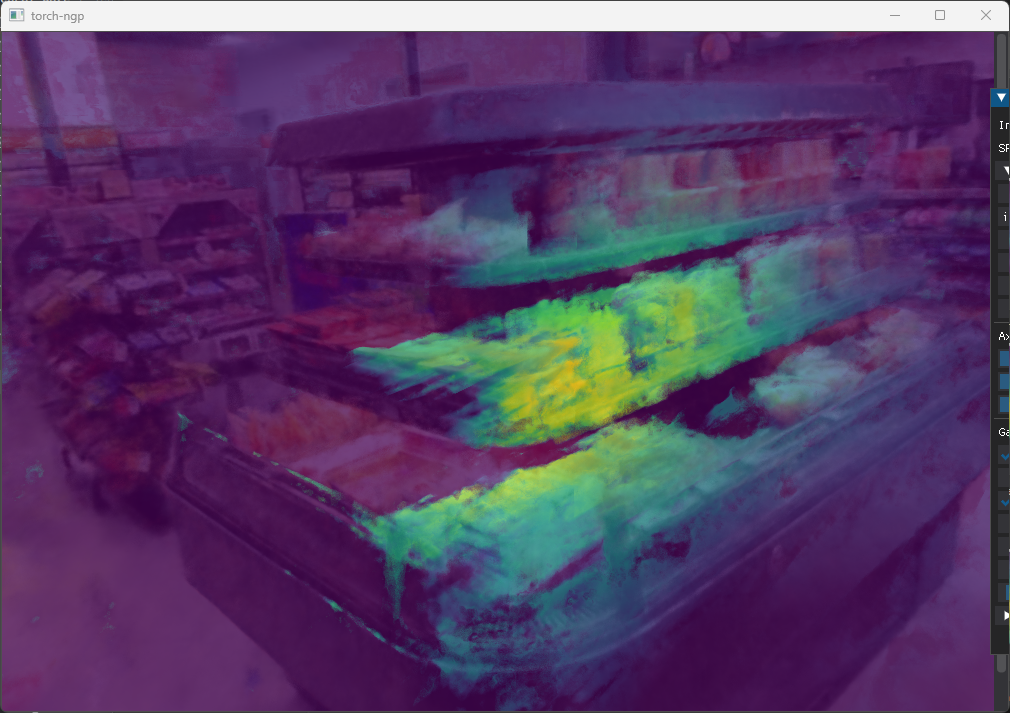}
         \caption{}
     \end{subfigure}
        \vspace{-0.25cm}
        \caption{
        Effect of gaze occlusion. (a) Scene rendered from the observer’s position depicting a store shelf. (b) Gaze predicted by NeRG. (c) Same observer as in (a), but the rendering camera is shifted left to look behind the shelf. \mbox{(d) As in (c)}, but with gaze occlusion applied. In (d), objects behind the shelf are not visible to the observer and therefore gaze is occluded; in (c), gaze propagates through the shelf.
        }
        \label{fig:shadow_exp}
\end{figure*}

\subsection{Visualizing Gaze with NeRG}

Analogous to our gaze probes, a trained NeRG model evaluates the gaze probability density at each queried ray. We visualize these values on 2D images as gaze heatmaps, with examples in \autoref{fig:gaze} comparing ground-truth and predicted gaze patterns. In \autoref{fig:shadow_exp}, we highlight the decoupling between the observer and the rendering camera, illustrating how NeRG handles gaze occlusion by modulating gaze based on surface visibility from the observer’s perspective; in \autoref{fig:shadow_exp}(d), the scene geometry occludes certain gaze rays, preventing the observer from attending to a surface that remains visible in the rendered image.

\section{Discussion}

\subsection{Comparison to Salience Mapping Methods}

Our method offers several advantages over existing techniques for 3D attention.
To begin, NeRG is explicitly informed by scene geometry through its use of a NeRF-based object representation, eliminating the need for separate models for depth estimation or 3D mapping of 2D salience. Unlike approaches that project 2D salience into 3D, NeRG directly predicts probability density values representing 3D gaze given an observer position and gaze ray direction, following the NeRF paradigm.

Moreover, we train NeRG on top of a fixed NeRF model, leaving the NeRF unchanged, and perform gaze prediction with a compact network based on the NeRF architecture, keeping the system lightweight and conceptually similar to NeRF.
Although NeRG is evaluated at every rendered pixel, its computational cost is comparable to NeRF and remains significantly lower than that of standard 2D salience models (under 100 ms per 720p frame on consumer-grade GPUs), enabling 3D gaze visualizations at interactive framerates. NeRG can also benefit from advances in NeRF-like models, offering potential for further speedups.

Unlike most prior work, our method also supports decoupling the observer from the rendering camera and accurately models gaze occlusion at inference, enabling finer user control and greater expressiveness in visualizing 3D gaze. For instance, in-store customer tracking data can be visualized from a bird’s-eye view to reveal, for any given observer position, which areas of the store attract customer attention. This decoupling is incompatible with 2D salience methods, which rely on image-space inputs that are difficult to generate in this configuration. 

Lastly, while our experiments use head pose as a proxy for gaze, NeRGs can also be adapted to use authentic gaze data or data from other sources, including 2D salience models. Although this extension may pose additional challenges, it is a promising direction for future work.

\subsection{Limitations}

\subsectionnonum{Training and evaluating NeRGs.} It is difficult to benchmark NeRG on existing 3D gaze data, because most gaze datasets are not designed for the training of NeRFs. In addition, these datasets commonly contain simple objects rather than complex 3D scenes where NeRGs excel. The alignment between gaze data and the NeRF-space coordinate system is also challenging. This step is, however, required to correctly represent gaze for rendered views of the scene. Training multiple NeRFs is also time consuming.
Lastly, most prior salience methods do not consider both complex 3D environments and gaze occlusion. We therefore struggled to define a fair comparison against existing techniques. 
We leave such metrics and benchmarks to future work.


\subsectionnonum{NeRF rendering artifacts.} Since NeRG is built on top of NeRFs, it naturally inherits their limitations. Rendering artifacts such as floaters can affect visualizations, and low-quality NeRF reconstructions can propagate to lower-quality NeRG outputs. The predicted gaze patterns may also exhibit their own artifacts or inaccuracies, because we model gaze with an additional NeRF-style network. Nonetheless, NeRG stands to benefit from continued advances in NeRFs, while improvements to NeRF reconstruction quality is outside the scope of this work.

\subsectionnonum{Accuracy of depth estimation.} NeRFs do not directly model depth, and the estimates of depth recovered from volume rendering may be inaccurate. However, NeRG uses the estimated depth to model the positions of the visible surfaces, as well as to test visibility between two spatial locations. Error in depth estimates can therefore result in inconsistent gaze prediction and gaze occlusion.

\subsectionnonum{Gaze occlusion.} While gaze occlusion is useful in visualizing gaze patterns for arbitrary observer perspectives decoupled from the rendering camera, authentic gaze data does not need to explicitly model it.  
Gaze tracking data naturally excludes gaze towards unseen surfaces: human observers simply do not pay attention to objects that they cannot see. 
In NeRG, gaze occlusion is only used when the the observer and rendering camera are decoupled, which isn't the case in the free viewing scenario. We therefore only evaluate the effect of gaze occlusion qualitatively.

\subsectionnonum{Using head pose to model gaze.} Our method assumes that head pose approximates the viewing direction and therefore gaze. While this assumption may be reasonable in some situations, e.g., large unstructured environments in natural outdoor settings, it may not hold true in structured environments like a store. For instance, head-tracking data from the convenience store may describe a person's movement through the store and not necessarily their attention.

\subsectionnonum{Using 2D salience to model gaze.} NeRGs can be trained on ground-truth gaze patterns predicted by 2D salience methods applied to views rendered by NeRF. However, 2D salience prediction algorithms may produce inaccurate salience maps in this particular application for two reasons. First, most state-of-the-art 2D salience prediction methods are trained on data acquired during free-viewing of 2D content (typically on a computer screen) \cite{mitSaliencyBenchmark, salicon2015}, which may be inadequate for 3D applications, especially when observers perform specific tasks. For instance, in a convenience store, shoppers are frequently engaged in search tasks, actively looking for products, which may require integrating 3D cues with semantic modeling. 


\section{Conclusion}

In this work, we presented Neural Radiance and Gaze Fields (NeRGs), a novel method for modeling and visualizing gaze patterns in complex 3D environments. By augmenting a pre-trained Neural Radiance Field (NeRF) with a gaze prediction module, NeRGs enable rendering of gaze patterns alongside scene reconstructions. Our approach supports gaze visualization from arbitrary observer viewpoints, optionally decoupled from the rendering camera, and considers gaze occlusion by 3D geometry to ensure spatially consistent gaze visualization.
We demonstrate the effectiveness of NeRGs through experiments in a real-world convenience store setting, leveraging head pose data from skeleton tracking to train our system.
Altogether, NeRGs enable real-time visualization of 3D gaze patterns from arbitrary observer and rendering perspectives, offering greater flexibility and user control in visualizing gaze in 3D scenes.

{
    \small
    \bibliographystyle{ieeenat_fullname}
    \bibliography{main}

\begin{thebibliography}{65}
\providecommand{\natexlab}[1]{#1}
\providecommand{\url}[1]{\texttt{#1}}
\expandafter\ifx\csname urlstyle\endcsname\relax
  \providecommand{\doi}[1]{doi: #1}\else
  \providecommand{\doi}{doi: \begingroup \urlstyle{rm}\Url}\fi

\bibitem[Adeli et~al.(2017)Adeli, Vitu, and Zelinsky]{adeli2017model}
Hossein Adeli, Fran{\c{c}}oise Vitu, and Gregory~J Zelinsky.
\newblock A model of the superior colliculus predicts fixation locations during scene viewing and visual search.
\newblock \emph{Journal of Neuroscience}, 37\penalty0 (6):\penalty0 1453--1467, 2017.

\bibitem[Barron et~al.(2021)Barron, Mildenhall, Tancik, Hedman, Martin-Brualla, and Srinivasan]{barron2021mip}
Jonathan~T Barron, Ben Mildenhall, Matthew Tancik, Peter Hedman, Ricardo Martin-Brualla, and Pratul~P Srinivasan.
\newblock {MiP-NeRF}: A multiscale representation for anti-aliasing neural radiance fields.
\newblock In \emph{Proceedings of the IEEE/CVF international conference on computer vision}, pages 5855--5864, 2021.

\bibitem[Barron et~al.(2022)Barron, Mildenhall, Verbin, Srinivasan, and Hedman]{barron2022mip360}
Jonathan~T. Barron, Ben Mildenhall, Dor Verbin, Pratul~P. Srinivasan, and Peter Hedman.
\newblock {Mip-NeRF 360}: Unbounded anti-aliased neural radiance fields.
\newblock In \emph{2022 IEEE/CVF Conference on Computer Vision and Pattern Recognition (CVPR)}, pages 5460--5469, 2022.

\bibitem[Barron et~al.(2023)Barron, Mildenhall, Verbin, Srinivasan, and Hedman]{barron2023zip}
Jonathan~T Barron, Ben Mildenhall, Dor Verbin, Pratul~P Srinivasan, and Peter Hedman.
\newblock {Zip-NeRF}: Anti-aliased grid-based neural radiance fields.
\newblock In \emph{Proceedings of the IEEE/CVF International Conference on Computer Vision}, pages 19697--19705, 2023.

\bibitem[Blender-Online-Community(2018)]{blender}
Blender-Online-Community.
\newblock \emph{Blender - a {3D} modelling and rendering package}.
\newblock Blender Foundation, Stichting Blender Foundation, Amsterdam, 2018.

\bibitem[Buswell(1935)]{buswellAttention1935}
Guy~Thomas Buswell.
\newblock \emph{How people look at pictures: a study of the psychology and perception in art.}
\newblock Univ. Chicago Press, Oxford, England, 1935.
\newblock Pages: 198.

\bibitem[Bylinskii et~al.()Bylinskii, Judd, Borji, Itti, Durand, Oliva, and Torralba]{mitSaliencyBenchmark}
Zoya Bylinskii, Tilke Judd, Ali Borji, Laurent Itti, Fr{\'e}do Durand, Aude Oliva, and Antonio Torralba.
\newblock {MIT} saliency benchmark.
\newblock http://saliency.mit.edu/.

\bibitem[Bylinskii et~al.(2019)Bylinskii, Judd, Oliva, Torralba, and Durand]{saliencyEvalMetrics}
Zoya Bylinskii, Tilke Judd, Aude Oliva, Antonio Torralba, and Fredo Durand.
\newblock What do different evaluation metrics tell us about saliency models?
\newblock \emph{IEEE Transactions on Pattern Analysis \& Machine Intelligence}, 41\penalty0 (03):\penalty0 740--757, 2019.

\bibitem[C2RO()]{C2RO}
C2RO.
\newblock https://www.c2ro.com/entera.

\bibitem[Cen et~al.(2023)Cen, Zhou, Fang, yang, Shen, Xie, Jiang, Zhang, and Tian]{segmentAnythingNerfs2023}
Jiazhong Cen, Zanwei Zhou, Jiemin Fang, chen yang, Wei Shen, Lingxi Xie, Dongsheng Jiang, Xiaopeng Zhang, and Qi Tian.
\newblock Segment anything in 3d with nerfs.
\newblock In \emph{Advances in Neural Information Processing Systems}, pages 25971--25990. Curran Associates, Inc., 2023.

\bibitem[Chong et~al.(2018)Chong, Ruiz, Wang, Zhang, Rozga, and Rehg]{chong2018connecting}
Eunji Chong, Nataniel Ruiz, Yongxin Wang, Yun Zhang, Agata Rozga, and James~M Rehg.
\newblock Connecting gaze, scene, and attention: Generalized attention estimation via joint modeling of gaze and scene saliency.
\newblock In \emph{Proceedings of the European conference on computer vision (ECCV)}, pages 383--398, 2018.

\bibitem[Chong et~al.(2020)Chong, Wang, Ruiz, and Rehg]{chong2020detecting}
Eunji Chong, Yongxin Wang, Nataniel Ruiz, and James~M Rehg.
\newblock Detecting attended visual targets in video.
\newblock In \emph{Proceedings of the IEEE/CVF conference on computer vision and pattern recognition}, pages 5396--5406, 2020.

\bibitem[Elmadjian et~al.(2018)Elmadjian, Shukla, Tula, and Morimoto]{elmadjian20183d}
Carlos Elmadjian, Pushkar Shukla, Antonio~Diaz Tula, and Carlos~H. Morimoto.
\newblock {3D} gaze estimation in the scene volume with a head-mounted eye tracker.
\newblock In \emph{Proceedings of the Workshop on Communication by Gaze Interaction}, New York, NY, USA, 2018. Association for Computing Machinery.

\bibitem[et. al.(2019)]{pytorch}
Adam~Paszke et. al.
\newblock {PyTorch}: An imperative style, high-performance deep learning library.
\newblock In \emph{Proceedings of the 33rd International Conference on Neural Information Processing Systems}, Red Hook, NY, USA, 2019. Curran Associates Inc.

\bibitem[Fisher(1953)]{vonmissesfisher}
Ronald Fisher.
\newblock Dispersion on a sphere.
\newblock \emph{Proceedings of the Royal Society of London. Series A, Mathematical and Physical Sciences}, 217\penalty0 (1130):\penalty0 295--305, 1953.

\bibitem[Foulsham et~al.(2011)Foulsham, Walker, and Kingstone]{gazeAllocation2011}
Tom Foulsham, Esther Walker, and Alan Kingstone.
\newblock The where, what and when of gaze allocation in the lab and the natural environment.
\newblock \emph{Vision Research}, 51\penalty0 (17):\penalty0 1920--1931, 2011.

\bibitem[Harel et~al.(2006)Harel, Koch, and Perona]{harel2006graph}
Jonathan Harel, Christof Koch, and Pietro Perona.
\newblock Graph-based visual saliency.
\newblock \emph{Advances in neural information processing systems}, 19, 2006.

\bibitem[Huang et~al.(2018)Huang, Cai, Li, and Sato]{huang2018predicting}
Yifei Huang, Minjie Cai, Zhenqiang Li, and Yoichi Sato.
\newblock Predicting gaze in egocentric video by learning task-dependent attention transition.
\newblock In \emph{Proceedings of the European conference on computer vision (ECCV)}, pages 754--769, 2018.

\bibitem[Huang et~al.(2020)Huang, Cai, and Sato]{huang2020ego}
Yifei Huang, Minjie Cai, and Yoichi Sato.
\newblock An ego-vision system for discovering human joint attention.
\newblock \emph{IEEE Transactions on Human-Machine Systems}, 50\penalty0 (4):\penalty0 306--316, 2020.

\bibitem[Itti et~al.(1998)Itti, Koch, and Niebur]{itti1998model}
Laurent Itti, Christof Koch, and Ernst Niebur.
\newblock A model of saliency-based visual attention for rapid scene analysis.
\newblock \emph{IEEE Transactions on pattern analysis and machine intelligence}, 20\penalty0 (11):\penalty0 1254--1259, 1998.

\bibitem[Jiang et~al.(2015)Jiang, Huang, Duan, and Zhao]{salicon2015}
Ming Jiang, Shengsheng Huang, Juanyong Duan, and Qi Zhao.
\newblock {SALICON}: Saliency in context.
\newblock In \emph{2015 IEEE Conference on Computer Vision and Pattern Recognition (CVPR)}, pages 1072--1080, 2015.

\bibitem[Johnson and Cuijpers(2013)]{johnson2013predicting}
David~O Johnson and Raymond~H Cuijpers.
\newblock Predicting gaze direction from head pose yaw and pitch.
\newblock In \emph{2013 International Conference on Image Processing, Computer Vision, \& Pattern Recognition (IPCV 2013)}, pages 662--668. World Academy Of Science, 2013.

\bibitem[Judd et~al.(2009)Judd, Ehinger, Durand, and Torralba]{judd2009learning}
Tilke Judd, Krista Ehinger, Fr{\'e}do Durand, and Antonio Torralba.
\newblock Learning to predict where humans look.
\newblock In \emph{2009 IEEE 12th international conference on computer vision}, pages 2106--2113. IEEE, 2009.

\bibitem[Kellnhofer et~al.(2019)Kellnhofer, Recasens, Stent, Matusik, and Torralba]{kellnhofer2019gaze360}
Petr Kellnhofer, Adria Recasens, Simon Stent, Wojciech Matusik, and Antonio Torralba.
\newblock Gaze360: Physically unconstrained gaze estimation in the wild.
\newblock In \emph{Proceedings of the IEEE/CVF international conference on computer vision}, pages 6912--6921, 2019.

\bibitem[Kingma and Ba(2015)]{adam}
Diederik~P. Kingma and Jimmy~Lei Ba.
\newblock Adam: A method for stochastic optimization.
\newblock In \emph{Proceedings of the International Conference on Learning Representations (ICLR)}, 2015.
\newblock Accepted papers -- Main Conference -- Poster Presentations.

\bibitem[Koch and Ullman(1987)]{koch1987shifts}
Christof Koch and Shimon Ullman.
\newblock Shifts in selective visual attention: towards the underlying neural circuitry.
\newblock In \emph{Matters of intelligence: Conceptual structures in cognitive neuroscience}, pages 115--141. Springer, 1987.

\bibitem[Krafka et~al.(2016)Krafka, Khosla, Kellnhofer, Kannan, Bhandarkar, Matusik, and Torralba]{krafka2016eye}
Kyle Krafka, Aditya Khosla, Petr Kellnhofer, Harini Kannan, Suchendra Bhandarkar, Wojciech Matusik, and Antonio Torralba.
\newblock Eye tracking for everyone.
\newblock In \emph{Proceedings of the IEEE conference on computer vision and pattern recognition}, pages 2176--2184, 2016.

\bibitem[Kruthiventi et~al.(2017)Kruthiventi, Ayush, and Babu]{kruthiventi2017deepfix}
Srinivas~SS Kruthiventi, Kumar Ayush, and R~Venkatesh Babu.
\newblock {DeepFix}: A fully convolutional neural network for predicting human eye fixations.
\newblock \emph{IEEE Transactions on Image Processing}, 26\penalty0 (9):\penalty0 4446--4456, 2017.

\bibitem[K{u}mmerer et~al.()K{u}mmerer, Wallis, and Bethge]{kummerersaliencyBenchmarking}
Matthias K{u}mmerer, Thomas S.~A. Wallis, and Matthias Bethge.
\newblock Saliency benchmarking made easy separating models, maps and metrics.
\newblock In \emph{Computer Vision – {ECCV} 2018}, pages 798--814. Springer International Publishing.

\bibitem[K{\"u}mmerer et~al.(2014)K{\"u}mmerer, Theis, and Bethge]{kummerer2014deep}
Matthias K{\"u}mmerer, Lucas Theis, and Matthias Bethge.
\newblock {Deep Gaze I}: Boosting saliency prediction with feature maps trained on {Imagenet}.
\newblock \emph{arXiv preprint arXiv:1411.1045}, 2014.

\bibitem[K{\"u}mmerer et~al.(2016)K{\"u}mmerer, Wallis, and Bethge]{kummerer2016deepgaze}
Matthias K{\"u}mmerer, Thomas~SA Wallis, and Matthias Bethge.
\newblock {DeepGaze II}: Reading fixations from deep features trained on object recognition.
\newblock \emph{arXiv preprint arXiv:1610.01563}, 2016.

\bibitem[Kummerer et~al.(2017)Kummerer, Wallis, Gatys, and Bethge]{kummerer2017understanding}
Matthias Kummerer, Thomas~SA Wallis, Leon~A Gatys, and Matthias Bethge.
\newblock Understanding low-and high-level contributions to fixation prediction.
\newblock In \emph{Proceedings of the IEEE international conference on computer vision}, pages 4789--4798, 2017.

\bibitem[K{\"u}mmerer et~al.(2022)K{\"u}mmerer, Bethge, and Wallis]{kummerer2022deepgaze}
Matthias K{\"u}mmerer, Matthias Bethge, and Thomas~SA Wallis.
\newblock {DeepGaze III}: Modeling free-viewing human scanpaths with deep learning.
\newblock \emph{Journal of Vision}, 22\penalty0 (5):\penalty0 7--7, 2022.

\bibitem[Kundu et~al.(2022)Kundu, Genova, Yin, Fathi, Pantofaru, Guibas, Tagliasacchi, Dellaert, and Funkhouser]{panopticNerf2022}
Abhijit Kundu, Kyle Genova, Xiaoqi Yin, Alireza Fathi, Caroline Pantofaru, Leonidas~J. Guibas, Andrea Tagliasacchi, Frank Dellaert, and Thomas Funkhouser.
\newblock Panoptic neural fields: A semantic object-aware neural scene representation.
\newblock In \emph{Proceedings of the IEEE/CVF Conference on Computer Vision and Pattern Recognition (CVPR)}, pages 12871--12881, 2022.

\bibitem[Lai et~al.(2023)Lai, Liu, Ryan, and Rehg]{lai2023eye}
Bolin Lai, Miao Liu, Fiona Ryan, and James~M Rehg.
\newblock In the eye of transformer: Global--local correlation for egocentric gaze estimation and beyond.
\newblock \emph{International Journal of Computer Vision}, pages 1--18, 2023.

\bibitem[Lavoué et~al.(2018)Lavoué, Cordier, Seo, and Larabi]{attention3Dshapes}
Guillaume Lavoué, Frédéric Cordier, Hyewon Seo, and Mohamed-Chaker Larabi.
\newblock Visual attention for rendered {3D} shapes.
\newblock \emph{Computer Graphics Forum}, 37\penalty0 (2):\penalty0 191--203, 2018.

\bibitem[Lee et~al.(2005)Lee, Varshney, and Jacobs]{lee2005mesh}
Chang~Ha Lee, Amitabh Varshney, and David~W Jacobs.
\newblock Mesh saliency.
\newblock In \emph{ACM SIGGRAPH 2005 Papers}, pages 659--666. ACM, 2005.

\bibitem[Li and Yu(2015)]{deepFeaturesSalience2015}
Guanbin Li and Yizhou Yu.
\newblock { Visual saliency based on multiscale deep features }.
\newblock In \emph{2015 IEEE Conference on Computer Vision and Pattern Recognition (CVPR)}, pages 5455--5463, Los Alamitos, CA, USA, 2015. IEEE Computer Society.

\bibitem[Li et~al.(2013)Li, Fathi, and Rehg]{Li_2013_ICCV}
Yin Li, Alireza Fathi, and James~M. Rehg.
\newblock Learning to predict gaze in egocentric video.
\newblock In \emph{Proceedings of the IEEE International Conference on Computer Vision (ICCV)}, 2013.

\bibitem[Linardos et~al.(2021)Linardos, K{\"u}mmerer, Press, and Bethge]{linardos2021deepgaze}
Akis Linardos, Matthias K{\"u}mmerer, Ori Press, and Matthias Bethge.
\newblock {DeepGaze IIE}: Calibrated prediction in and out-of-domain for state-of-the-art saliency modeling.
\newblock In \emph{Proceedings of the IEEE/CVF International Conference on Computer Vision}, pages 12919--12928, 2021.

\bibitem[Liu et~al.(2023{\natexlab{a}})Liu, Zhang, Zheng, and Duan]{semanticRay2023}
Fangfu Liu, Chubin Zhang, Yu Zheng, and Yueqi Duan.
\newblock Semantic ray: Learning a generalizable semantic field with cross-reprojection attention.
\newblock In \emph{2023 IEEE/CVF Conference on Computer Vision and Pattern Recognition (CVPR)}, pages 17386--17396, Los Alamitos, CA, USA, 2023{\natexlab{a}}. IEEE Computer Society.

\bibitem[Liu et~al.(2023{\natexlab{b}})Liu, Hu, Huang, Tai, and Tang]{instanceNerf2023}
Yichen Liu, Benran Hu, Junkai Huang, Yu-Wing Tai, and Chi-Keung Tang.
\newblock Instance neural radiance field.
\newblock In \emph{2023 IEEE/CVF International Conference on Computer Vision (ICCV)}, pages 787--796, Los Alamitos, CA, USA, 2023{\natexlab{b}}. IEEE Computer Society.

\bibitem[Lou et~al.(2022)Lou, Lin, Marshall, Saupe, and Liu]{TranSalNet2022}
Jianxun Lou, Hanhe Lin, David Marshall, Dietmar Saupe, and Hantao Liu.
\newblock {TranSalNet}: Towards perceptually relevant visual saliency prediction.
\newblock \emph{Neurocomputing}, 494:\penalty0 455--467, 2022.

\bibitem[MacInnes et~al.(2018)MacInnes, Iqbal, Pearson, and Johnson]{macinnes2018wearable}
Jeff~J MacInnes, Shariq Iqbal, John Pearson, and Elizabeth~N Johnson.
\newblock Wearable eye-tracking for research: Automated dynamic gaze mapping and accuracy/precision comparisons across devices.
\newblock \emph{BioRxiv}, page 299925, 2018.

\bibitem[Martin et~al.(2024)Martin, Fandos, Masia, and Serrano]{mesh3DSaliency2024}
Daniel Martin, Andres Fandos, Belen Masia, and Ana Serrano.
\newblock {SAL3D}: a model for saliency prediction in {3D} meshes.
\newblock \emph{Vis. Comput.}, 40\penalty0 (11):\penalty0 7761–7771, 2024.

\bibitem[Maurus et~al.(2014)Maurus, Hammer, and Beyerer]{maurus2014realistic}
Michael Maurus, Jan~Hendrik Hammer, and J{\"u}rgen Beyerer.
\newblock Realistic heatmap visualization for interactive analysis of {3D} gaze data.
\newblock In \emph{Proceedings of the Symposium on Eye Tracking Research and Applications}, pages 295--298, 2014.

\bibitem[Mildenhall et~al.(2021)Mildenhall, Srinivasan, Tancik, Barron, Ramamoorthi, and Ng]{mildenhall2021nerf}
Ben Mildenhall, Pratul~P Srinivasan, Matthew Tancik, Jonathan~T Barron, Ravi Ramamoorthi, and Ren Ng.
\newblock {NeRF}: Representing scenes as neural radiance fields for view synthesis.
\newblock \emph{Communications of the {ACM}}, 65\penalty0 (1):\penalty0 99--106, 2021.

\bibitem[M\"uller et~al.(2022)M\"uller, Evans, Schied, and Keller]{mueller2022instantNgp}
Thomas M\"uller, Alex Evans, Christoph Schied, and Alexander Keller.
\newblock Instant neural graphics primitives with a multiresolution hash encoding.
\newblock \emph{ACM Trans. Graph.}, 41\penalty0 (4):\penalty0 102:1--102:15, 2022.

\bibitem[Naas et~al.(2020)Naas, Jiang, Sigg, and Ji]{naas2020functional}
Si-Ahmed Naas, Xiaolan Jiang, Stephan Sigg, and Yusheng Ji.
\newblock Functional gaze prediction in egocentric video.
\newblock In \emph{Proceedings of the 18th International Conference on Advances in Mobile Computing \& Multimedia}, pages 40--47, 2020.

\bibitem[Nonaka et~al.(2022)Nonaka, Nobuhara, and Nishino]{nonaka2022dynamic}
Soma Nonaka, Shohei Nobuhara, and Ko Nishino.
\newblock Dynamic {3D} gaze from afar: Deep gaze estimation from temporal eye-head-body coordination.
\newblock In \emph{Proceedings of the IEEE/CVF Conference on Computer Vision and Pattern Recognition}, pages 2192--2201, 2022.

\bibitem[Song et~al.(2019)Song, Liu, and Rosin]{song2019}
Ran Song, Yonghuai Liu, and Paul Rosin.
\newblock Mesh saliency via weakly supervised classification-for-saliency cnn.
\newblock \emph{IEEE Transactions on Visualization and Computer Graphics}, PP:\penalty0 1--1, 2019.

\bibitem[Song et~al.(2023)Song, Zhang, Zhao, Liu, and Rosin]{song20233d}
Ran Song, Wei Zhang, Yitian Zhao, Yonghuai Liu, and Paul~L Rosin.
\newblock {3D} visual saliency: An independent perceptual measure or a derivative of 2d image saliency?
\newblock \emph{IEEE Transactions on Pattern Analysis and Machine Intelligence}, 45\penalty0 (11):\penalty0 13083--13099, 2023.

\bibitem[Tavakoli et~al.(2019)Tavakoli, Rahtu, Kannala, and Borji]{tavakoli2019digging}
Hamed~Rezazadegan Tavakoli, Esa Rahtu, Juho Kannala, and Ali Borji.
\newblock Digging deeper into egocentric gaze prediction.
\newblock In \emph{2019 IEEE Winter Conference on Applications of Computer Vision (WACV)}, pages 273--282. IEEE, 2019.

\bibitem[Thakur et~al.(2021)Thakur, Beyan, Morerio, and Del~Bue]{thakur2021predicting}
Sanket~Kumar Thakur, Cigdem Beyan, Pietro Morerio, and Alessio Del~Bue.
\newblock Predicting gaze from egocentric social interaction videos and imu data.
\newblock In \emph{Proceedings of the 2021 International Conference on Multimodal Interaction}, pages 717--722, 2021.

\bibitem[Vig et~al.(2014)Vig, Dorr, and Cox]{vig2014large}
Eleonora Vig, Michael Dorr, and David Cox.
\newblock Large-scale optimization of hierarchical features for saliency prediction in natural images.
\newblock In \emph{Proceedings of the IEEE conference on computer vision and pattern recognition}, pages 2798--2805, 2014.

\bibitem[Vora et~al.(2022)Vora, Radwan, Greff, Meyer, Genova, Sajjadi, Pot, Tagliasacchi, and Duckworth]{nesf2022}
Suhani Vora, Noha Radwan, Klaus Greff, Henning Meyer, Kyle Genova, Mehdi S.~M. Sajjadi, Etienne Pot, Andrea Tagliasacchi, and Daniel Duckworth.
\newblock Ne{SF}: Neural semantic fields for generalizable semantic segmentation of {3D} scenes.
\newblock \emph{Transactions on Machine Learning Research}, 2022.

\bibitem[Wang et~al.(2018)Wang, Pi, Qin, Shen, and Shi]{wang2018slam}
Haofei Wang, Jimin Pi, Tong Qin, Shaojie Shen, and Bertram~E Shi.
\newblock {SLAM}-based localization of {3D} gaze using a mobile eye tracker.
\newblock In \emph{Proceedings of the 2018 ACM symposium on eye tracking research \& applications}, pages 1--5, 2018.

\bibitem[Wang et~al.(2024)Wang, Panchadsaram, Sherkati, and Clark]{Wang2024eyeTracking}
Yinan Wang, Sansitha Panchadsaram, Rezvan Sherkati, and James~J. Clark.
\newblock An egocentric video and eye-tracking dataset for visual search in convenience stores.
\newblock \emph{Computer Vision and Image Understanding}, 248:\penalty0 104129, 2024.

\bibitem[Xu et~al.(2024)Xu, Cheng, Gao, Wang, Gao, and Shan]{xu2024instantmesh}
Jiale Xu, Weihao Cheng, Yiming Gao, Xintao Wang, Shenghua Gao, and Ying Shan.
\newblock Instantmesh: Efficient {3D} mesh generation from a single image with sparse-view large reconstruction models.
\newblock \emph{arXiv preprint arXiv:2404.07191}, 2024.

\bibitem[Yang and Li(2021)]{yang2021visual}
Bincheng Yang and Hongwei Li.
\newblock A visual attention model based on eye tracking in {3D} scene maps.
\newblock \emph{ISPRS International Journal of Geo-Information}, 10\penalty0 (10):\penalty0 664, 2021.

\bibitem[Ye et~al.(2012)Ye, Li, Fathi, Han, Rozga, Abowd, and Rehg]{ye2012detecting}
Zhefan Ye, Yin Li, Alireza Fathi, Yi Han, Agata Rozga, Gregory~D Abowd, and James~M Rehg.
\newblock Detecting eye contact using wearable eye-tracking glasses.
\newblock In \emph{Proceedings of the 2012 ACM conference on ubiquitous computing}, pages 699--704, 2012.

\bibitem[Zhang and Sclaroff(2013)]{zhang2013saliency}
Jianming Zhang and Stan Sclaroff.
\newblock Saliency detection: A boolean map approach.
\newblock In \emph{Proceedings of the IEEE international conference on computer vision}, pages 153--160, 2013.

\bibitem[Zhang et~al.(2008)Zhang, Tong, Marks, Shan, and Cottrell]{zhang2008sun}
Lingyun Zhang, Matthew~H Tong, Tim~K Marks, Honghao Shan, and Garrison~W Cottrell.
\newblock {SUN}: A bayesian framework for saliency using natural statistics.
\newblock \emph{Journal of vision}, 8\penalty0 (7):\penalty0 32--32, 2008.

\bibitem[Zhang et~al.(2017)Zhang, Teck~Ma, Hwee~Lim, Zhao, and Feng]{zhang2017deep}
Mengmi Zhang, Keng Teck~Ma, Joo Hwee~Lim, Qi Zhao, and Jiashi Feng.
\newblock Deep future gaze: Gaze anticipation on egocentric videos using adversarial networks.
\newblock In \emph{Proceedings of the IEEE conference on computer vision and pattern recognition}, pages 4372--4381, 2017.

\bibitem[Zhi et~al.(2021)Zhi, Laidlow, Leutenegger, and Davison]{semanticNerf2021}
Shuaifeng Zhi, Tristan Laidlow, Stefan Leutenegger, and Andrew~J. Davison.
\newblock In-place scene labelling and understanding with implicit scene representation.
\newblock In \emph{2021 IEEE/CVF International Conference on Computer Vision (ICCV)}, pages 15818--15827, 2021.

\end{thebibliography}
}

\end{document}